\documentclass[12pt]{article}

\usepackage{times}
\usepackage[a4paper, margin=1in]{geometry}
\usepackage{enumerate}
\usepackage{amsmath, amssymb, amsthm,amsfonts,color}
\usepackage{hyperref}  
\usepackage{parskip}
\usepackage{verbatim}
\usepackage{upgreek}

\usepackage{epsfig,mathrsfs,graphicx,euscript,amsxtra}

\theoremstyle{remark}

\numberwithin{equation}{section}

\begin{document}
\title{U-Net with ResNet Backbone for Garment Landmarking Purpose}
\author{Kelvin Hong Khay Boon\footnote{This work is done during employment in Saratix, Custlr in Malaysia.}  \\ email \color{blue}{\href{mailto:kh.boon2@gmail.com}{kh.boon2@gmail.com}}}
\date{15 Feb 2022}

\maketitle

\begin{abstract}
We build a heatmap-based landmark detection model to locate important landmarks on 2D RGB garment images.
The main goal is to detect edges, corners and suitable interior region of the garments. 
This let us re-create 3D garments in modern 3D editing software by incorporate landmark detection model and texture unwrapping. 
We use a U-net architecture with ResNet backbone to build the model. With an appropriate loss function, we are able to train a moderately robust model.
\end{abstract}

\section{Introduction}
\begin{figure}[ht]
    \centering
    \includegraphics[width=12cm]{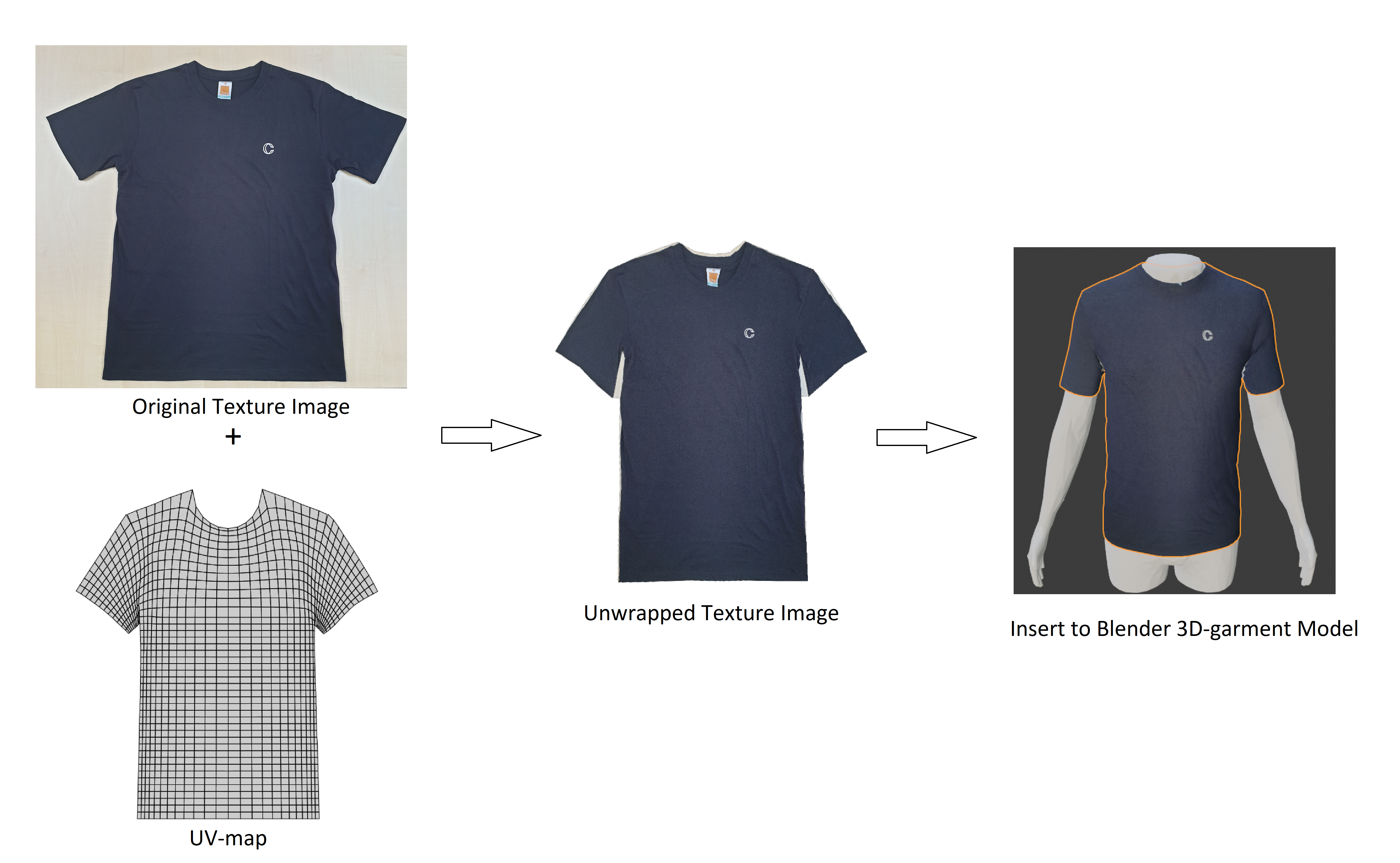}
    \caption{Logic of transfer garment texture to 3D-garment.}
    \label{Pipeline}
\end{figure}

In SaratiX, one of our project is to provide 3D-garment reconstruction from 2D-images. 
Specifically, we take a garment's front and back images, then try to reconstruct it in a 3D virtual environment.
Currently, we must know the garment type beforehand, then construct a template mesh for each garment type to be prepared for texture unwrapping. 
With this in mind, we develop a landmark detection model to find the contour of the garment and important parts of it, such as collar and armpit. 
Then, we use these information to perform geometric image transformation on the texture images so it will follows the UVmaps of our template 3D-garments (Figure \ref{Pipeline}). 

We use a deep learning approach to detect important landmarks on various type of garments. 
In this article, we choose T-shirt for illustration purpose. 
We adopt U-net architecture with Residual neural network as our backbone (PyTorch Pretrained Resnet34 Model) to accelerate training.

\section{Setup}
\begin{figure}[ht]
    \centering
    \includegraphics[width=10cm]{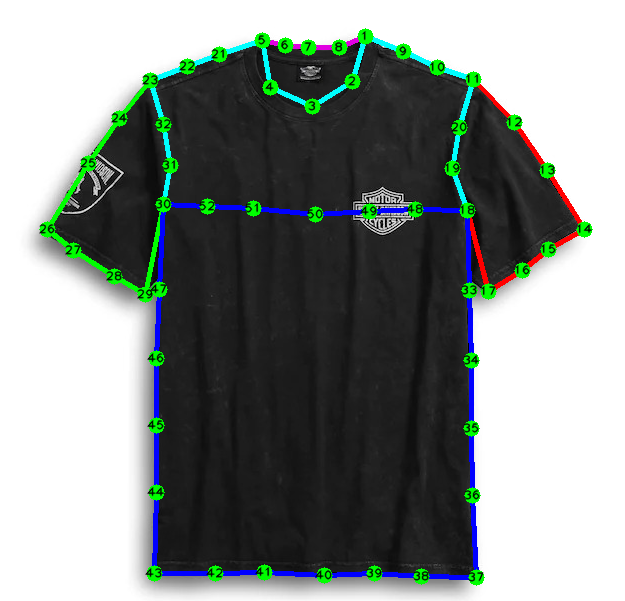}
    \caption{Landmark design on T-shirt}
    \label{tshirt_landmark}
\end{figure}

We take T-shirt as an example, then label 52 points on it.
In Figure \ref{tshirt_landmark}, the line segments are only for cosmetic purpose, illustrating the contours of the T-shirt.

We prepare 250 images to train the landmarks on T-shirt, and the result looks good after training for 400 epoches (we will discuss the loss function soon).

\section{Model Architecture}

\begin{figure}[ht]
    \centering
    \includegraphics[width=15cm]{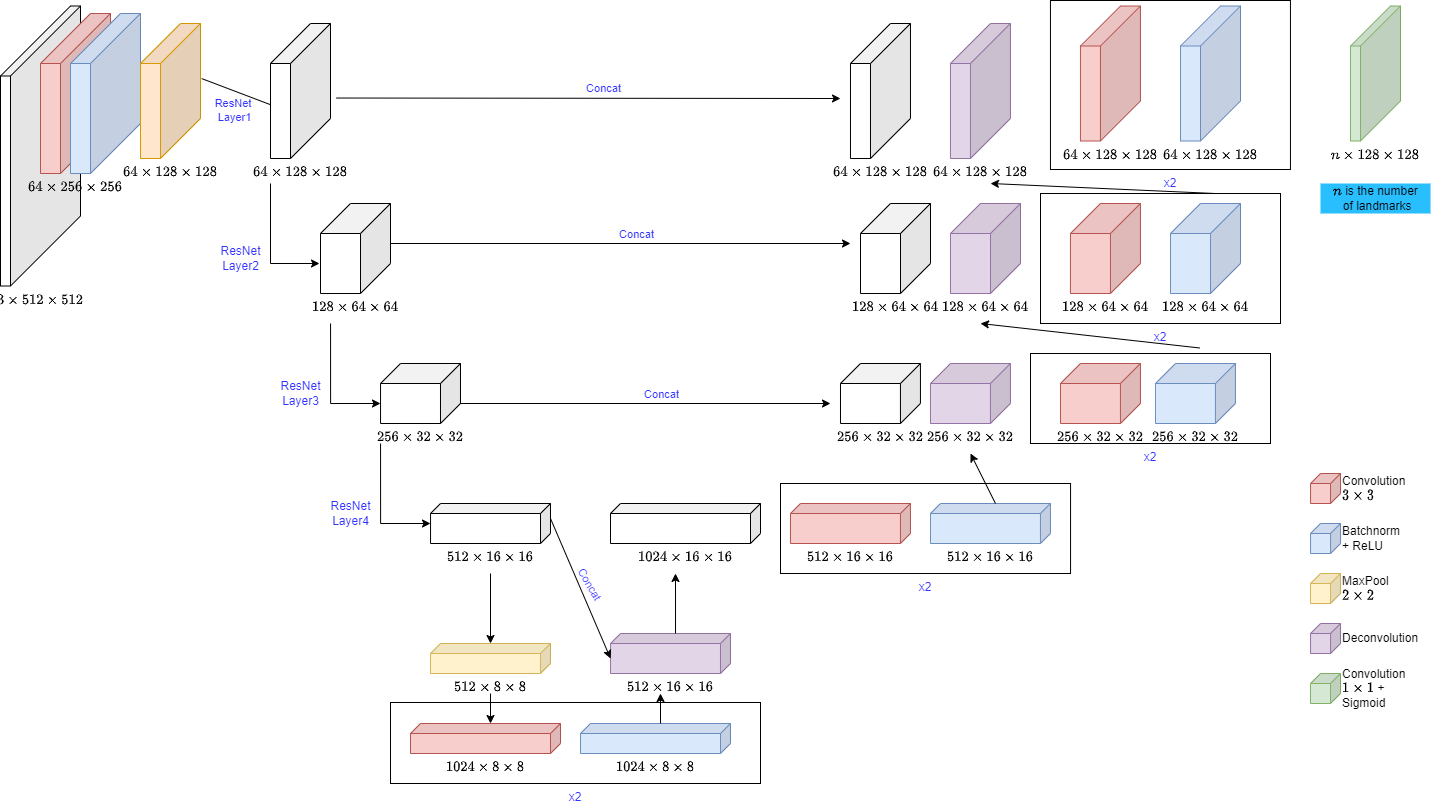}
    \caption{Resnet-U-Net architecture}
    \label{model_architecture}
\end{figure}

We use a U-net architecture with pretrained Resnet34 weights (from PyTorch) as a backbone to build our heatmap-based landmarking model (see Figure \ref{model_architecture}). 
Using a single RGB image input, we first resize the image to $512\times 512$. 
Then, the model will output a tensor of shape $n\times 128\times 128$ responsible for storing the heatmaps, where $n$ stands for the number of landmarks we need.

\begin{figure}[ht]
    \centering
    \includegraphics[width=15cm]{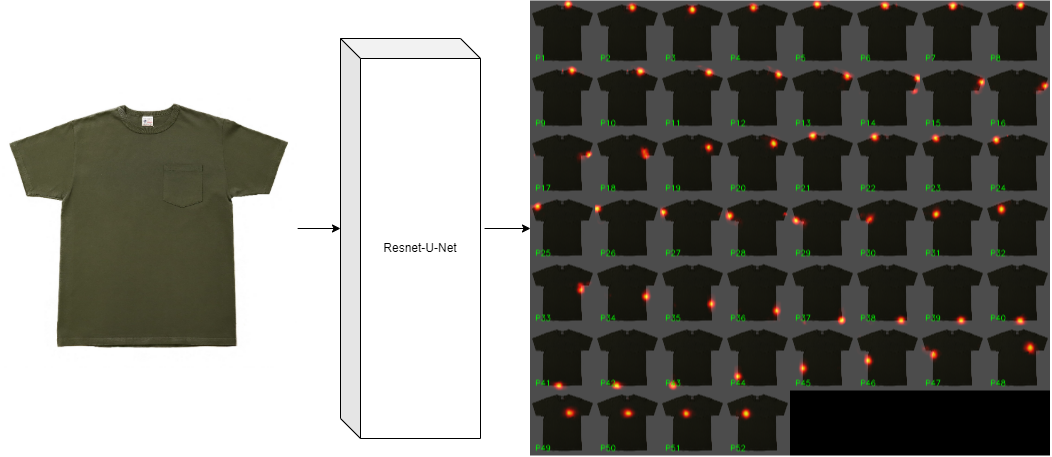}
    \caption{Output visualization}
    \label{visualization}
\end{figure}

\begin{figure}[ht]
    \centering
    \includegraphics[width=13cm]{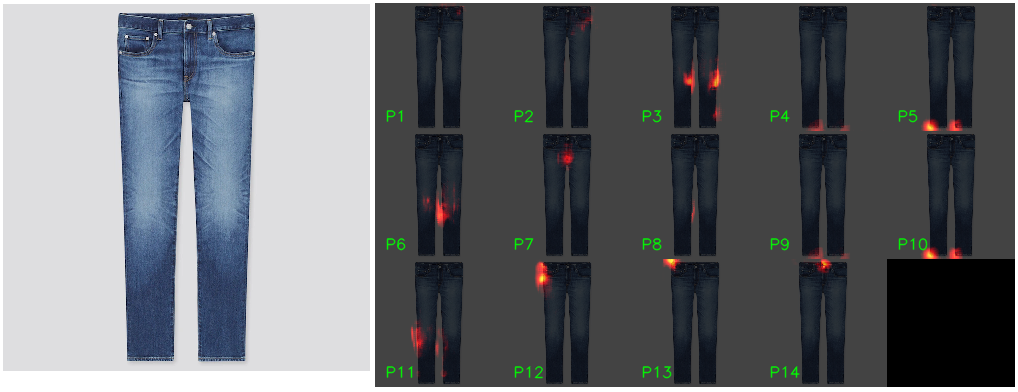}
    \caption{Double Attentions on P3, P5 and P10.}
    \label{double_attention}
\end{figure}

Below (Figure \ref{visualization}) shows a visualization of the output, in the case of 52 landmarks on T-shirt. 
We use a $7\times 8$ grid to show the heatmap stack on top of the original T-shirt image. 
Our goal is to train the model for heatmap detection. 
We choose heatmap over regression because it simply works better in our case.
On top of that, the heatmap outputs give us plenty of insights about the performance of our model compare to regression method, such as avoiding double attentions (Figure \ref{double_attention}). 
Each heatmap is a $128\times 128$ matrix with real entries in $[0,1]$. 
Pixel with value near 0 represent black color, while it is represented by red color around $0.5$, and then yellow color around $1$. 


\begin{figure}[ht]
    \centering
    \includegraphics[width=8cm]{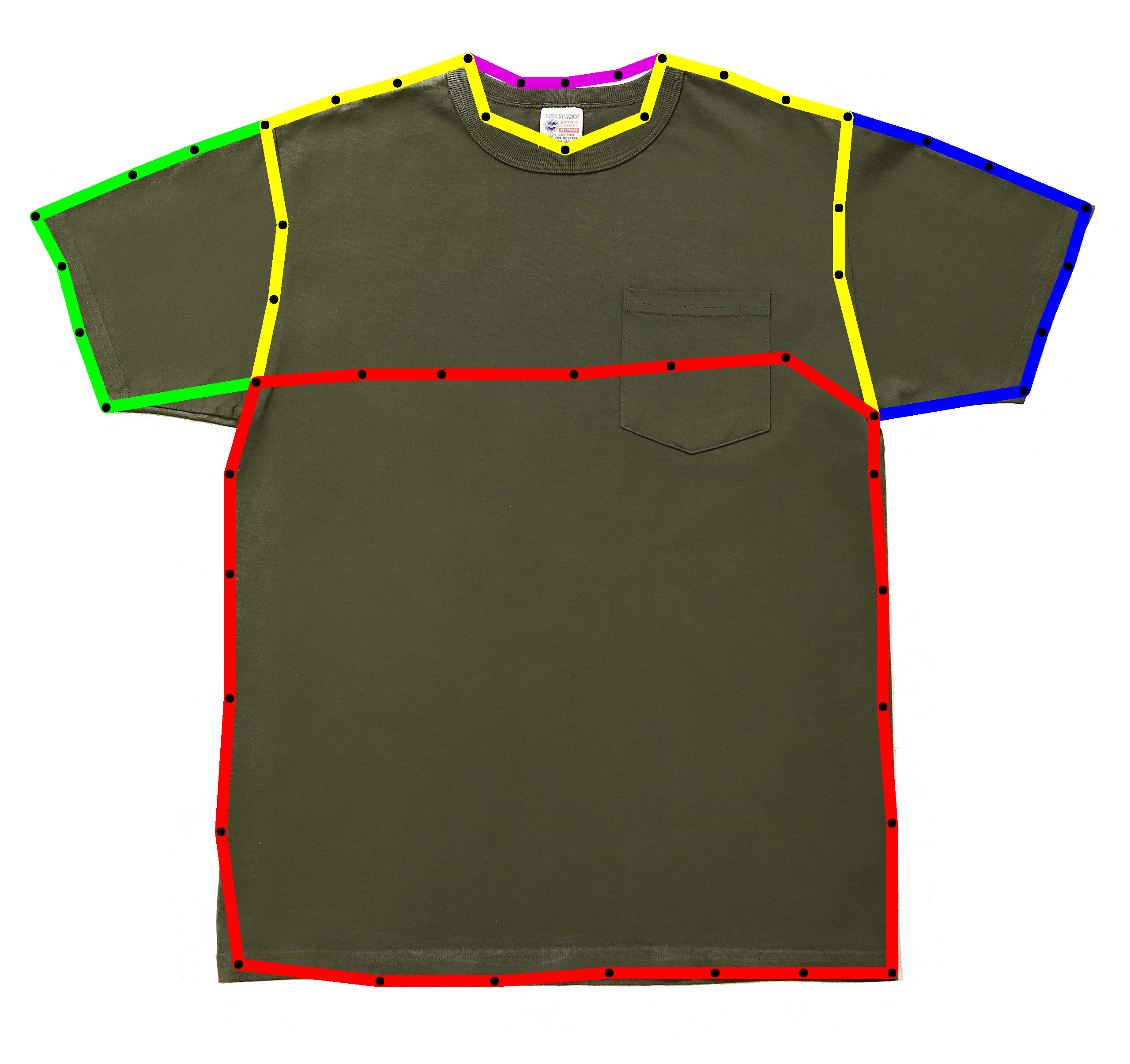}
    \caption{Showing landmarks on T-shirt}
    \label{tshirt_output}
\end{figure}

After getting the $n$ heatmaps (Figure \ref{tshirt_output}), we calculate the index of maximum pixel to get the landmark coordinates. 

\pagebreak

\section{Dataset and Loss function}
Before training, we did augmentation for the dataset, and this augmentation is being done on the fly.
To train a robust model, we did random shear and random rotation on the dataset. 

Since we are using heatmap approach for landmark detection, we use a weighted loss function to make the model equally emphasize on the narrow landmark location and the comparably vast background region. 
Let $n$ be the total number of landmarks, then the model will output $n$ heatmaps $H_1, \dots, H_n$ where each $H_i$ is a matrix of size $128\times 128$. 
Let the ground truth coordinates be $a_1, \dots, a_n$, where each $a_i=(x_i,y_i)$ is the location of the landmark on the image. 
We will assume the landmarks are integers and have been resized to the region of $128\times 128$, which means $0\leq x_i\leq 127, 0\leq y_i\leq 127$. 

Let $r=10$ be the radius of our landmarks. 
We transform the set of landmarks $a_1,\dots,a_n$ to be the set of ground truth heatmaps $G_1,\dots, G_n$, where
$G_i$ satisfies the following: The $(x_i,y_i)$th entry of $G_i$ is 1, and for those entry with distance from $a_i$ at least $r$ pixel equals 0. 
We then set $G_i$ to be interpolated linearly for those pixels with distance from $a_i$ less than 10. 
This is to mimic the heatmap representation of each landmark. 

With the purpose of training $\{H_i\}$ towards $\{G_i\}$, we purposely added Sigmoid function as the last layer of our model.

We could use a standard regression loss function: 
$$\frac1n\sum_{i=1}^n \dfrac{\|H_i-G_i\|}{128^2}$$
where $\|A\|=\sum_{i,j} |A_{ij}|$ is the L1 norm for any matrix $A$, to train our model. 
However, as background consists of majority of the heatmaps, the model will tend to predict heatmaps that are entirely black. 
Therefore, we need to weight the component of loss function. 
We define the indicator matrix $I_i$ of $G_i$ such that 
$$(I_i)_{jk} = \begin{cases}1&\text{ if }(G_i)_{jk} > 0,\\
0&\text{ if }(G_i)_{jk} = 0.\end{cases}$$
In other words, $I_i$ is a binary mask that is white on the disk of radius 10 around $a_i$, black on the remaining pixels. 
Moreover, we also define element-wise product $Z$ of two matrices $X,Y$ of same shape as
$$Z=X\odot Y$$
where $Z_{ij} = X_{ij}Y_{ij}$. 

We can now formally define our loss function as follows: Given ground truth landmarks $G = \{a_1,\dots,a_n\}$, we can get the associated ground truth heatmaps $\{G_1,\dots,G_n\}$ and their indicators $\{I_1, \dots, I_n\}$. For the heatmap predictions $P = \{H_1,\dots,H_n\}$, the loss is defined by
$$L(P, G) = \frac1n\sum_{i=1}^n \left[ \frac12\cdot \dfrac{\|I_i\odot (H_i-G_i)\|}{\|I_i\|}+\frac12\cdot \dfrac{\|(1-I_i)\odot (H_i-G_i)\|}{\|1-I_i\|}\right],$$
where $1$ is recognized as the matrix of ones with shape $128\times 128$. 
In this function, different weights are distributed to solve the unbalance issue from the background. 
Notice the normalization factors are used to make the value of the loss function falls between 0 and 1, enable us to judge the performance of the model easily. 
For example, if the loss is $0.2$, we can conclude that the pixel values of the predictions differ $0.2$ averagely from the ground truths.

\section{Training}
With a batch size of 8, learning rate $\lambda=0.005$, we use Adam optimizer to train our model for 400 epoches. 
We use $80\%$ of our dataset for training, and the remaining $20\%$ for validation. 

\begin{figure}[ht]
    \centering
    \includegraphics[width=8cm]{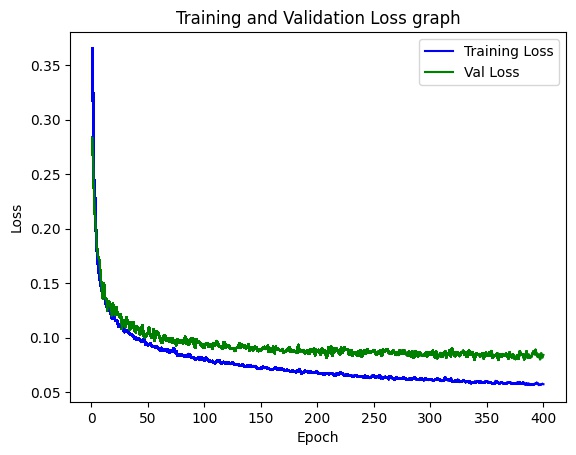}
    \caption{Losses}
    \label{losses}
\end{figure}

Figure \ref{losses} shows a graph of training and validation loss. 

\pagebreak
\section{Discussion}
\begin{figure}[ht]
    \centering
    \includegraphics[width=14cm]{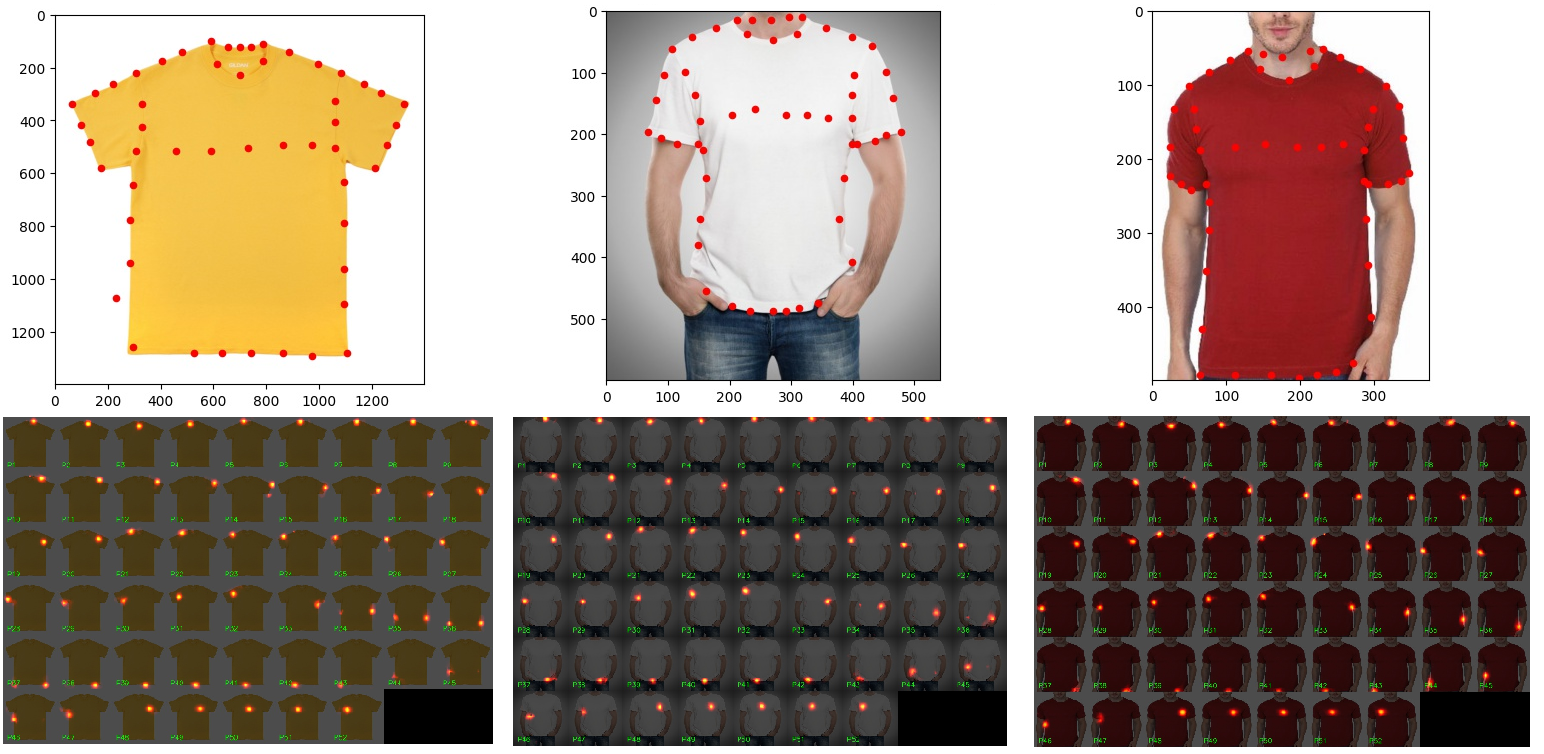}
    \caption{Test on unseen data}
    \label{example}
\end{figure}

We use the model to perform inferencing on some unseen images (Figure \ref{example}).
There are still flaws within our model.
For example, the yellow shirt on the left has some unstable prediction on the bottom-left corner. 
On the other two images, the model couldn't predict occluded landmarks behind their left hands. 
We conjecture these are because the last layer of our model is too simple, and probably need some extension from that point. 

There are some methods which could possibly improve the model.
First, we could use Active Shape Model (similar to \cite{asmnet}) to better stabilize the predictions, so the result of the yellow T-shirt above would not happen; it could also detect occluded landmarks since the machine knows the overall contour of the landmarks. 
Furthermore, we could also add some hidden layers before the last layer to interchange information among the $n$ heatmaps, so that nearby landmarks can relate to each other, possibly improve the predictions. 

Finally, our dataset is still small (currently 200 images for each model). 
The model could highly benefit by simply increase the dataset to 500 images. 

\section{Acknowledgments}
The outcome of this deep learning model, datasets and the result from this paper is thanks to the support of the leaders and colleagues in SaratiX, Custlr Sdn. Bhd.
Moreover, I also gained excellent knowledge on how to build deep neural networks in Recogine Sdn. Bhd. during my previous internship period. 
This work certainly cannot be done in such intergrity without the help of supportive and genius colleagues from these companies.

\end{document}